\documentclass[runningheads, anonymous=true]{llncs}
\usepackage[T1]{fontenc}
\usepackage{graphicx}
\usepackage{dblfloatfix}
\begin{document}
\title{Unimodal and Multimodal Representation Training for Relation Extraction}
%
%
\author{Ciaran Cooney\ \and
Rachel Heyburn\ \and 
Liam Madigan\ \and 
Mairead O'Cuinn\ \and 
Chloe Thompson\ \and 
Joana Cavadas}

\institute{Aflac NI, Belfast, Northern Ireland \\
\email{\{ccooney,rheyburn,lmadigan,mocuinn,cthompson,jcavadas\}@aflac.com}}
\authorrunning{C. Cooney et al.}
%
%
\maketitle              
\begin{abstract}
Multimodal integration of text, layout and visual information has achieved SOTA results in visually rich document understanding (VrDU) tasks, including relation extraction (RE). However, despite its importance, evaluation of the relative predictive capacity of these modalities is less prevalent. Here, we demonstrate the value of shared representations for RE tasks by conducting experiments in which each data type is iteratively excluded during training. In addition, text and layout data are evaluated in isolation. While a bimodal text and layout approach performs best (F1=0.684), we show that text is the most important single predictor of entity relations. Additionally, layout geometry is highly predictive and may even be a feasible unimodal approach. Despite being less effective, we highlight circumstances where visual information can bolster performance. In total, our results demonstrate the efficacy of training joint representations for RE.

\end{abstract}
\section{Introduction}
With many sectors such as healthcare, insurance and e-commerce now relying on digitization and artificial intelligence to exploit document information, Visually-rich Document Understanding (VrDU) has become a highly active research domain~\cite{zhang2020trie,liu2019graph,xu2020layoutlmv2,jaume2019funsd}. VrDU is the task of analyzing scanned or digital business documents to allow structured information to be extracted for downstream business applications \cite{xu2020layoutlmv2}. Sub-fields including Named-Entity Recognition (NER)~\cite{carbonell2020neural}, layout understanding~\cite{DBLP:journals/corr/abs-2003-02356} and document classification~\cite{xu2020layoutlm} all seek to extract meaningful information from documents. Another sub-field of VrDU, relation extraction (RE) offers the possibility of linking named entities in documents so that a paired relationship can be identified~\cite{jaume2019funsd,davis2021visual,dang2021end,carbonell2021named,xu2021layoutxlm}. Typically, relations are defined in a question-answer (\emph{Q/A}) format and the RE task is to define a function which predicts if a pair of entities in a document are related or not~\cite{jaume2019funsd,xu2021layoutxlm}.  

Concurrent with recent developments in VrDU, advances in multimodal deep learning have seen novel methods applied across fields as diverse as medical imaging~\cite{sharif2021decision}, neurotechnology~\cite{cooney2021bimodal} and early prediction of Alzheimer's disease~\cite{venugopalan2021multimodal}. Commercial and opensource optical character recognition engines such as AWS Textract\footnote{\label{note2}\url{https://aws.amazon.com/textract/}}, Microsoft Read API\footnote{\label{note3}\url{https://docs.microsoft.com/en-us/azure/cognitive-services/computer-vision/overview-ocr}} and PyTesserect\footnote{\label{note4}\url{https://pypi.org/project/pytesseract/}} enable extraction of detailed text and geometric information from visually-rich documents, which along with visual information has led to a plethora of multimodal architectures being applied to VrDU tasks~\cite{liu2019graph,li2021structurallm,wei2020robust,zhang2020trie,xu2020layoutlm,xu2020layoutlmv2}. These approaches enable learning of joint representations in a single end-to-end training procedure with the aim of maximising the total information in a document. Although transformer-based architectures are prominent in this field~\cite{xu2021layoutxlm,li2021structext}, other methods for optimizing RE tasks, such as graph neural networks~\cite{carbonell2021named,davis2021visual}, have been reported.

Datasets for RE facilitate the use of multimodal representations for training co-adaptive networks~\cite{jaume2019funsd,xu2021layoutxlm}. However, despite the growth of multimodal approaches in VrDU tasks, the extent to which learning joint representations is an enhancement remains unclear, as does the relative capacity of each data type. Although text is likely more predictive than either geometric layout or visual information, the extent to which this is the case and the interaction between modes is not known. Even in studies with ablation tests, consideration of the effects of training without text representations has not been applied~\cite{hong2021bros}.

We aim to address uncertainty regarding the predictive capacity of different data by performing a series of experiments with different multimodal and unimodal configurations. We apply our analysis to the RE task, due to it being an unresolved information extraction challenge relevant to several industry applications, and one that could benefit from appropriately trained joint representations. Our contributions  are summarised as follows: (1) We prove the efficacy of using joint representations for VrDU RE. Specifically we demonstrate that a text/layout configuration yields the best performance. (2)	We analyse the asymmetric predictive capacity of text, layout and visual data, exhibiting the anticipated relative importance of text over the other data while highlighting where layout and visual information can be effectual. (3) We present a simplified classifier for RE based on the LayoutXLM classification head~\cite{xu2021layoutxlm}. 

Section 2 describes previous works related to multimodal approaches to document understanding tasks and questions current understanding of the impact of different modalities. Section 3 reports our methodology, including the dataset used, model architecture, and experimental procedures. In Section 4, we present the results of our experiments. Section 5 contains limitations and suggestions for future work. In Section 6 we forward concluding remarks.

\section{Related Work}
Two datasets exist for the VrDU RE task. These are FUNSD \cite{jaume2019funsd} and XFUND \cite{xu2021layoutxlm}. Both contain annotations which include an indication of linked entities comprising of two entity IDs which are linked, or an empty array indicating no relationship.

The provision of text, geometry and  document images in these datasets enables the use of multimodal methods for document understanding.  The LayoutLM family of models~\cite{xu2020layoutlm,xu2020layoutlmv2,xu2021layoutxlm} utilise combined text and position embeddings to leverage the layout of the document, with LayoutLMv2 extending this approach to fully incorporate visual information. The approach of~\cite{wang2020docstruct} uses a similar trifecta of inputs to perform tasks on the FUNSD and MedForm datasets.~\cite{audebert2019multimodal} use a multimodal approach for text and image-based document classification, while others focus on text and layout representations~\cite{pramanik2020towards,li2021structurallm,li2021structext}.

The use of multimodal approaches poses an important question: \textit{what are the relative effects of the different data types?} It is not always clear from reported results, even in studies that do present ablation findings, what the impact of different modalities is. Particularly since text since is usually retained in conducted experiments~\cite{pramanik2020towards,hong2021bros}. For industry applications, the additional training and inference costs associated with large-scale multimodal approaches must be mitigated by performance benefits. The original XFUND paper does not report on the relative impact of the different components (text, layout and visual information)~\cite{xu2021layoutxlm}, informing the approach taken here.

\section{Methodology}

We use the XFUND dataset\footnote{\label{note1}\url{https://github.com/doc-analysis/XFUND}} to experiment with different modalities for the RE task~\cite{xu2021layoutxlm}. The dataset consists of document images for form understanding in seven languages. Annotations corresponding to each of the documents contain a unique identifier, class label, bounding box coordinates (\( {x_{left},y_{top},x_{right},y_{bottom}}\)), text and a linking indicator. This linking indicator facilitates the use of XFUND in VrDU RE. Entities for RE are designated key-value pairs corresponding to questions and answers in the forms. For further information on dataset collection and curation, see~\cite{xu2021layoutxlm}. Dataset statistics differ from those reported in~\cite{xu2021layoutxlm} and are therefore presented in Table~\ref{tab:xfun stats}. \emph{ZH}, \emph{JA}, \emph{ES}, \emph{FR}, \emph{IT}, \emph{DE} and \emph{PT} correspond to Chinese, Japanese, Spanish, French, Italian, German and Portuguese, respectively.

\begin{table}
  \caption{Train/Test split for XFUND data.}
  \begin{center}
  \begin{tabular}{|c|l|l|l|l|l|l|l|}
    \hline
    {} & \verb|ZH| & \verb|JA| & \verb|ES| & \verb|FR| & \verb|IT| & \verb|DE| & \verb|PT|\\
    \hline
    Train & {187} & {194} & {243} & {202} & 265 & 189 & 233\\
    \hline
    Test & \centering {65} & {71} & {74} & {71} & 92 & 63 & 85\\
    \hline
\end{tabular}
\end{center}
\label{tab:xfun stats}
\end{table}

\begin{figure*}[h]
  \centering
  \includegraphics[width=\textwidth]{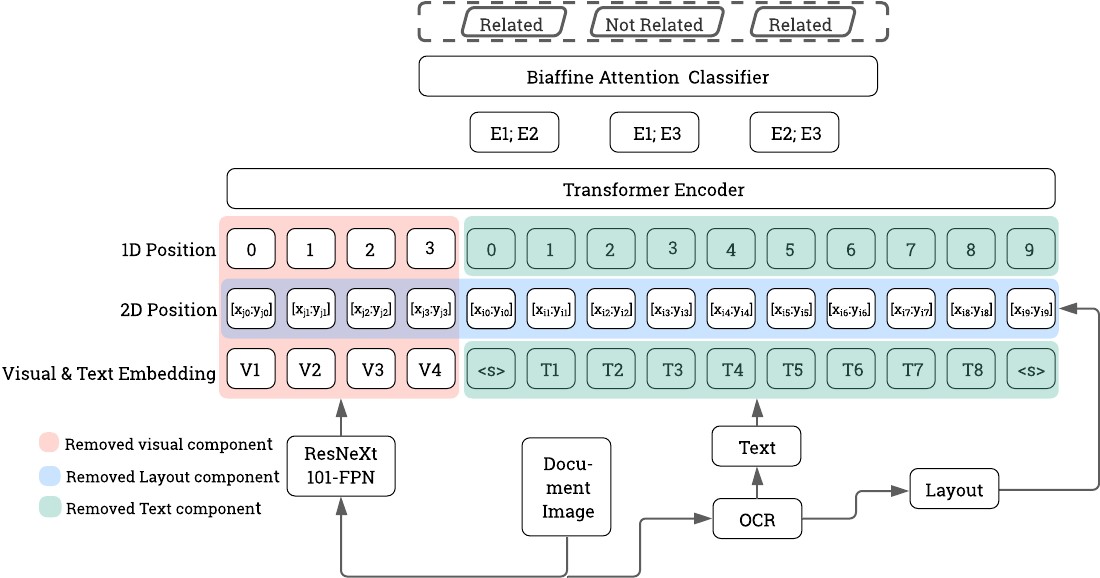}
  \caption{Multimodal transformer with data exlusions color-coded. Pink denotes exclusion of viusal components, blue exclusion of layout, and green exclusion of text representations.}
  \label{fig:transformer}
\end{figure*}

\subsection{LayoutXLM for Relation Extraction}
\label{subsection: layoutxlm}

The multimodal deep learning architecture we use to perform our experiments is LayoutXLM, a pretrained transformer for document understanding~\cite{xu2021layoutxlm}, based on the LayoutLMv2 architecture~\cite{xu2020layoutlmv2}. The model ingests text, layout (bounding boxes) and visual information which are encoded in embedding layers (Figure~\ref{fig:transformer}). It utilises traditional position embeddings to model word position in a sequence, and 2D-position embeddings to consider relative spatial position. A visual backbone encodes image representations using ResNeXt 101-FPN~\cite{Xie2016}.

For the RE task, a bespoke classification layer is attached to the pretrained model for further fine-tuning. In the original LayoutXLM, a bi-affine classifier receives representations of \emph{Q/A} entities which are a processed version of the first token vector and an entity type embedding for each. A feed-forward neural network is applied to these representations before they are fed to the bi-affine classifer. Here, we have slightly modified the classification layer to simplify the feed-forward neural network and therefore reduce the number of parameters. In this work the feed-forward neural network consists of a single fully-connected layer, leaky relu activation and dropout (\emph{p}=0.2).

\subsection{Experimental Procedures}
\label{subsection: experiments}
We conduct experiments for six different multimodal and unimodal configurations for each of the seven sets in the multilingual XFUND data. Experiments consist of fine-tuning the pretrained LayoutXLM model on various configurations of the available data. The six model configurations are: (1) Multimodal text, layout and visual (MM), (2) bimodal text and layout (text/layout), (3) bimodal text and visual (text/visual), (4) bimodal layout and visual (layout/visual), (5) unimodal layout (layout) and (6) unimodal text (text). We initially planned to include a unimodal visual experiment but early results indicated this was not feasible for the RE task.

For experiment 1, there are no further modifications to the network beyond those specified in Section~\ref{subsection: layoutxlm}. For experiment 2, all visual components of the architecture are removed. This includes the visual backbone and all related embeddings, including 2D and 1D visual position embeddings  (Figure~\ref{fig:transformer}; pink). The model is therefore only trained on text and layout information. In experiment 3, layout information is removed from the network in the form of 2D position embeddings. Included in this step is the removal of 2D position embeddings from the visual component (Figure~\ref{fig:transformer}; blue). Experiment 4 excludes all text information, including tokenized text and the associated 1D and 2D position embeddings (Figure~\ref{fig:transformer}; green). Tasks 5 and 6 combine the relevant exclusions applied in experiments 2, 3 and 4. Studies often neglect to experiment with the exclusion of text~\cite{pramanik2020towards,hong2021bros}, despite its importance when analysing the predictive capacity of multimodal approaches.

Due to the nature of the ablation experiments we hypothesised that a degree of variation in optimal learning rates across the different data types and perhaps between the individual datasets was likely. For this reason, learning rate is the only hyperparameter optimized. Three learning rates were optimized with simple gridsearch: 5e$^{-5}$, 1e$^{-5}$ and 5e$^{-6}$. All other learning parameters are identical across experiments. For fine-tuning, batch size is 2, and all models are allowed to train for 50 epochs.

\section{Results}
For all experiments, F1 score is the primary evaluation metric, with precision and recall also reported. Although multimodal results have previously been reported for the XFUND RE task~\cite{xu2021layoutxlm}, we chose not to include them in our reporting due differences in network configuration, dataset statistics and training procedures.

\subsection{Bimodal training outperforms trimodal}
\begin{figure}[t]
  \centering
  \includegraphics[width=\linewidth]{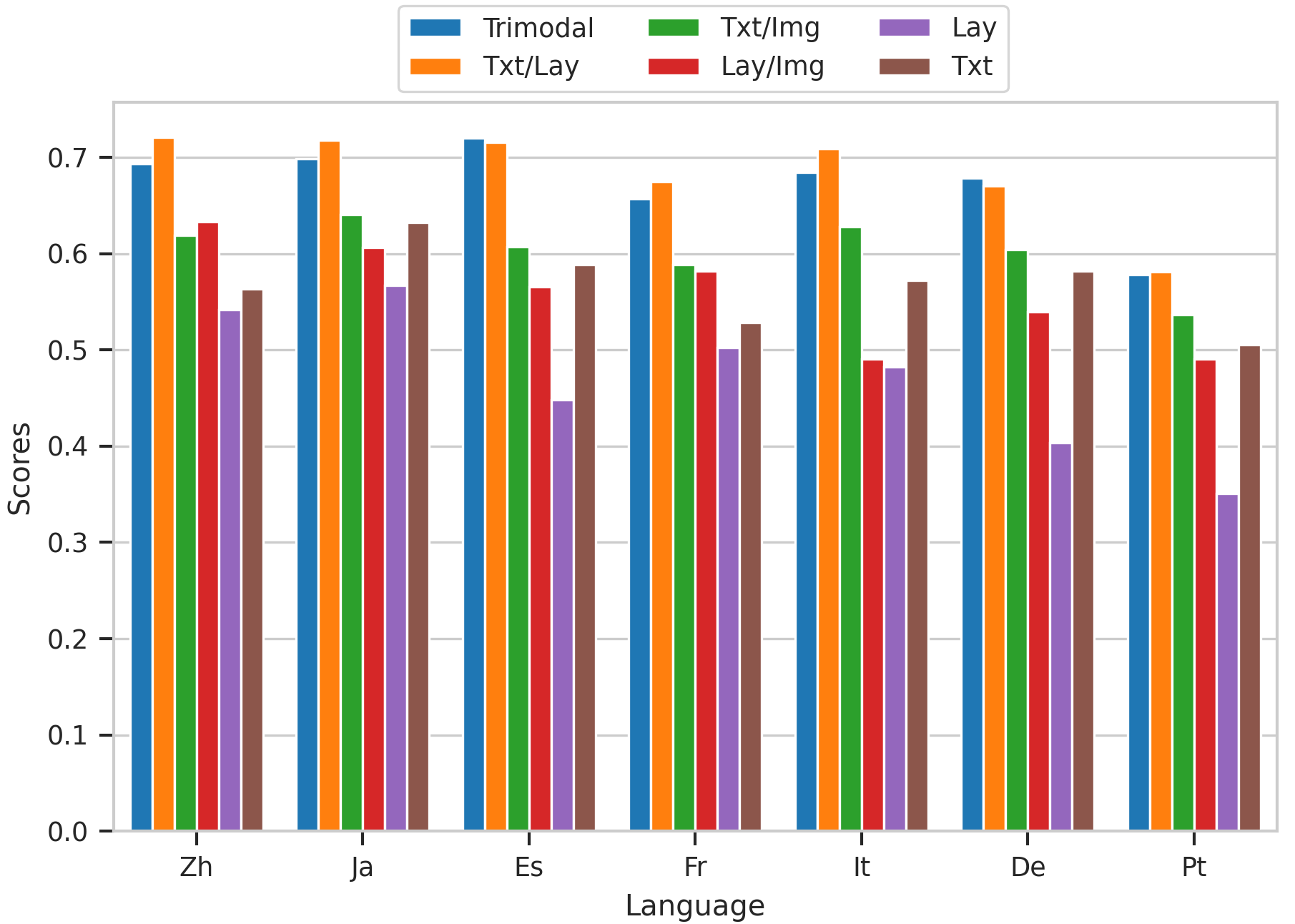}
  \caption{Bar chart reporting the validation F1 scores for each language dataset with each network configuration.}
  \label{fig:barchart}
\end{figure}

F1 scores obtained from models trained with each of the six different network configurations are reported in Table~\ref{tab:xfun results} and Figure~\ref{fig:barchart}. Additional recall and precision scores are in Tables~\ref{tab:xfun recall} and~\ref{tab:xfun precision}. Overall results validate the utility of training on joint representations for the VrDU RE task, while also exhibiting asymmetric predictive capacity across the different modalities (Table~\ref{fig:barchart}). Here, the bimodal text/layout configuration outperforms the three-pronged multimodal approach with mean F1 scores of 0.684 and 0.673, respectively. This is consistent with previous work demonstrating the utility of text and layout information without the requirement of visual information~\cite{li2021structurallm}. These two training configurations significantly outperform all other approaches with scores 8.08\% and 6.93\% greater than the next best approach (Figure~\ref{fig:barchart}). The other bimodal approaches, text/visual and layout/visual, result in reasonably strong F1 scores of 0.604 and 0.558 and further suggest that joint representations can be effective for this task. However, the overall picture indicates that in this particular application of multimodal deep learning there is a hierarchy of predictive capacity with text on top and visual information at the bottom.

Pairwise comparison of results from full multimodal training and the text/
layout configuration indicate that the impact of visual information is negligible or that its inclusion is even counterproductive (Table~\ref{tab:xfun results}). However, comparison of unimodal and bimodal results suggest visual information can be effective in the correct context. Most obvious is the positive impact of visual information when compared to the unimodal layout results. In this case the inclusion of visual data improves upon layout only results by improving the F1 score from 0.471 to 0.558. In fact, the combination of these two modalities results in similar performance to the unimodal text method. It is possible that this approach is useful in scenarios where text quality is degraded but visual and layout information are sufficient to classify the document. However, results do not suggest that visual information is necessary for the RE task.

Significant variation in performance corresponding to the different document languages may have been expected, particularly as there is a clear dichotomy between those using Kanji characters and those using the Latin alphabet. Despite some of the latin languages exhibiting less effective impact of layout information, Figure~{\ref{fig:barchart}} indicates similar levels of overall performances across datasets, despite differences in the number of document samples per set (Table~{\ref{tab:xfun stats}}).

\begin{table}
  \caption{XFUND F1 scores for different training configurations.}
  \begin{center}
  \begin{tabular}{|c|l|l|l|l|l|l|l|}

    \hline
    {}&MM&Txt/Lay & Txt/Im & Lay/Im & Layout & Text \\
    \hline
    \verb|ZH| & {0.6935} & \textbf{0.7212} & {0.6192} & {0.6334} & 0.5417 & 0.5636\\
    \hline
    \verb|JA| & {0.6987} & \textbf{0.7181} & {0.6406} & {0.6061} & 0.5674 & 0.6321\\
    \hline
    \verb|ES| & \textbf{0.7198} & {0.7159} & {0.6069} & {0.5657} & 0.4483 & 0.5885\\
    \hline
    \verb|FR| & {0.6573} & \textbf{0.6747} & {0.5888} & {0.5820} & 0.5021 & 0.5285\\
    \hline
    \verb|IT| & {0.6841} & \textbf{0.7090} & {0.6281} & {0.4906} & 0.4825 & 0.5724\\
    \hline
    \verb|DE| & \textbf{0.6782} & {0.6701} & {0.6041} & {0.5397} & 0.4035 & 0.5821\\
    \hline
    \verb|PT| & {0.5779} & \textbf{0.5812} & {0.5367} & {0.4909} & 0.3511 & 0.5053\\
    \hline
  Mean & 0.6728 & \textbf{0.6843} & 0.6035 & 0.5583 & 0.4709 & 0.5675 \\
  
  \hline
\end{tabular}
\end{center}
\label{tab:xfun results}
\end{table}

\begin{table}
  \caption{Recall scores for XFUND data for different training configurations.}
  \begin{center}
  \begin{tabular}{|c|l|l|l|l|l|l|l|}
    \hline
    {}&MM&Txt/Lay & Txt/Im & Lay/Im & Layout & Text \\
    \hline
    \verb|ZH| & {0.6109} & \textbf{0.7607} & {0.6754} & {0.7011} & 0.6639 & 0.6440\\
    \hline
    \verb|JA| & {0.5638} & \textbf{0.7540} & {0.6601} & {0.6619} & 0.6932 & 0.6762\\
    \hline
    \verb|ES| & {0.6475} & \textbf{0.7210} & {0.6817} & {0.6807} & 0.4487 & 0.6136\\ 
    \hline
    \verb|FR| & {0.6231} & \textbf{0.7278} & {0.6365} & {0.7278} & 0.6956 & 0.5774\\
    \hline
    \verb|IT| & {0.6314} & \textbf{0.7090} & {0.6584} & {0.5649} & 0.5966 & 0.7009\\
    \hline
    \verb|DE| & \textbf{0.6990} & {0.6518} & {0.6267} & {0.6158} & 0.4279 & 0.5931\\
    \hline
    \verb|PT| & {0.4851} & \textbf{0.6640} & {0.5308} & {0.5891} & 0.4082 & 0.5089\\
    \hline
    Mean & 0.6087 & \textbf{0.7126} & 0.6385 & 0.6488 & 0.5620 & 0.6163 \\
    \hline
\end{tabular}
\end{center}
\label{tab:xfun recall}
\end{table}

\begin{table}[!b]
  \caption{Precision scores for XFUND data for different training configurations.}
  \begin{center}
  \begin{tabular}{|c|l|l|l|l|l|l|l|}
    \hline
    {}&MM&Txt/Lay & Txt/Im & Lay/Im & Layout & Text \\
    \hline
    \verb|ZH| & {0.6109} & \textbf{0.6855} & {0.5717} & {0.5777} & 0.4576 & 0.5010\\
    \hline
    \verb|JA| & {0.5638} & \textbf{0.6854} & {0.6223} & {0.5589} & 0.4802 & 0.5934\\
    \hline
    \verb|ES| & {0.6475} & \textbf{0.7108} & {0.5469} & {0.4768} & 0.4478 & 0.5654\\
    \hline
    \verb|FR| & {0.6231} & \textbf{0.6288} & {0.5478} & {0.4848} & 0.3928 & 0.4872\\
    \hline
    \verb|IT| & {0.6314} & \textbf{0.6862} & {0.6004} & {0.4336} & 0.4050 & 0.4837\\
    \hline
    \verb|DE| & \textbf{0.6990} & {0.6894} & {0.5831} & {0.5804} & 0.3817 & 0.5715\\
    \hline
    \verb|PT| & {0.4851} & \textbf{0.5167} & {0.5427} & {0.4207} & 0.3080 & 0.5016\\
    \hline
    Mean & 0.6087 & \textbf{0.6575} & 0.5736 & 0.5046 & 0.4104 & 0.5291 \\
    \hline
\end{tabular}
\end{center}
\label{tab:xfun precision}
\end{table}

\subsection{Text is the anchor for relation extraction}

As expected, results clearly indicate the most significant drop-off in F1 score occurs when text is excluded. Of the six approaches, the two for which text is excluded are the poorest performing. The fact that the unimodal text method exhibits better classification performance than the bimodal layout/visual approach is a strong indicator of the dominance of text for the RE task.

Notwithstanding, the inclusion of supplementary data to enhance performance is extremely valuable. Results show that each of the three configurations that included the addition of other modalities alongside text produced improved performance. In this case, text is necessary but not sufficient to achieve the best possible performance. Clearly, layout information provides very important supplementary information for the RE task. Not only is it used along with text in the highest performing configuration but it also exhibits reasonable performance in the absence of text, in both unimodal (0.471) and bimodal (0.558) networks.

Despite clear dominance, there is notable variation in the strength of predictions across languages in the absence of text representations. For ZH and FR there is minimal difference or even improvement when text is replaced with layout information, whereas for IT and DE this difference is substantial (Table~\ref{tab:xfun results}). This suggests that document diversity across regions or business sectors may modulate the effectiveness of different approaches to training joint representations. It may not always be obvious \emph{a priori} whether or not supplementing text data with other modalities will actually enhance model performance. Given trade-offs associated with speed and complexity, particularly with the inclusion of visual information, efforts should be made to evaluate the value proposition when applying multimodal techniques in information retrieval tasks.

\begin{figure}[t]
  \centering
  \includegraphics[width=\linewidth]{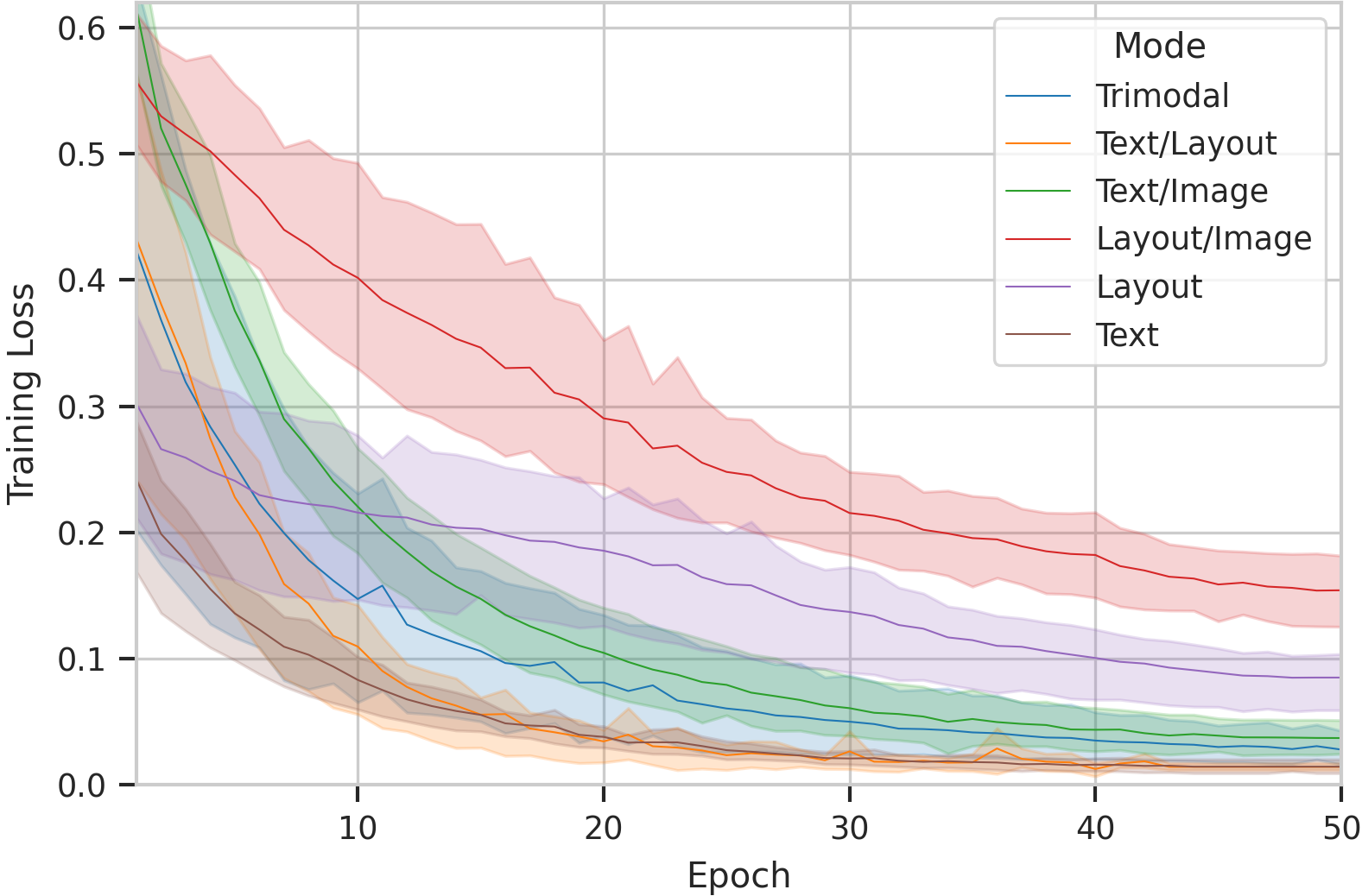}
  \caption{Training loss over 50 epochs for each network configuration. Shadows indicate the range of loss across each language dataset for a given epoch.}
  \label{fig:linechart}
\end{figure}

\subsection{Training variability is data dependent}

The model's receptivity to learning the RE task from text is further illuminated by training loss trends for each configuration (Figure~\ref{fig:linechart}). A much sharper decline towards convergence is present in those model configurations considering text than those excluding it. Each of the four approaches using text converge on a similar loss after 50 epochs. The other two methods (layout and layout/image) exhibit shallower learning trajectories, greater variance between datasets and less ability to converge within the prescribed training time. Again this validates the extent to which text is the most important anchor in training even when other data modalities are also included. The higher variance in loss exhibited by the non-text approaches may also indicate that the effectiveness of layout and image data may be more dependent on specific datasets than text. Another factor influencing these results is the optimal learning rate selected for each experiment. While this varied somewhat between the different language sets within experiments, there is more obvious variation between experiments. The full multimodal approach and the text/layout approach are both trained with an optimal learning rate of 5e$^{-5}$. Experiments with text excluded used learning rates of either 1e$^{-5}$ or 5e$^{-6}$.

\section{Limitations and Future Work}

Limitations associated with this work include the relative sparsity of data currently available for the VrDU RE task and the constraint of applying the experiments to LayoutXLM only. To the best of our knowledge FUNSD \cite{jaume2019funsd} is the only other dataset currently available for the RE task. Increasing the diversity of datasets and also the volume of samples within datasets could allow us to validate these results further. Additionally, other multimodal architectures exist with different approaches to encoding and combining modalities \cite{gu2022xylayoutlm,huang2022layoutlmv3}. Extending this work to include analysis of these different methods would provide a stronger basis on which to judge the relative contributions of text, layout and visual information to the RE task.

As well as addressing these limitations, future work may involve a large-scale analysis of multimodal approaches to a variety of VrDU tasks. This could include document classification, semantic entity recognition, and key information extraction tasks and ablation analysis could be applied to a variety of model architectures in order to fully understand how their performance differs according to modality. In addition, time and complexity analysis is required to understand the relative utility of different approaches within business environments that may process documents at high daily volumes. This would facilitate understanding of the efficiency and cost value of the different methods.

\section{Conclusions}
Multimodal methods can have trade-offs with respect to complexity, performance and speed when applied to industry applications. We trained a multimodal transformer using several data configurations to understand the impact of training joint representations for the VrDU RE task. The bimodal text and layout approach resulted in the best performance, even beating the full multimodal configuration. Individually, text accounts for a greater portion of the predictive capacity than either layout or visual data. Unimodal text achieved higher mean F1 score than bimodal layout/visual data and all configurations with text included outperformed those with text excluded. Nevertheless, both layout and visual information was proven to be effective in specific conditions. Although, visual information is currently exhibiting value as a supplementary data source to boost overall performance, our results show that layout information is extremely important to the RE task. We also showed that training various depending on the inclusion/exclusion of different data types. Future work is required to examine methods for optimizing joint representation training for document understanding, including how to best combine different data in multimodal approaches.
%
%

%
%
%
\bibliographystyle{splncs04}
%
\bibliography{bibliography}



\end{document}